\documentclass[letterpaper, 10pt, conference]{ieeeconf}

\IEEEoverridecommandlockouts                              

\usepackage[T1]{fontenc}
\usepackage{graphics}   
\usepackage{graphicx}
\usepackage{epsfig}     
\usepackage{times}      
\usepackage{amsmath}    
\usepackage{amssymb}    
\usepackage{amsfonts}
\usepackage{subcaption}
\usepackage{textcomp}
\usepackage{blindtext}
\usepackage{booktabs}
\usepackage{algorithm,algpseudocode}
\usepackage{url}
\usepackage{hyperref}
\usepackage{cite}
\usepackage{balance}

\makeatletter
\newcounter{phase}[algorithm]
\newlength{\phaserulewidth}
\newcommand{\setphaserulewidth}{\setlength{\phaserulewidth}}
\newcommand{\phase}[2]{%
  \vspace{-1ex}
  \Statex\leavevmode\llap{\rule{\dimexpr\labelwidth+\labelsep}{\phaserulewidth}}\rule{\linewidth}{\phaserulewidth}
  \Statex\strut\refstepcounter{phase}\textit{Stage~#1 -- #2}
  \vspace{-1ex}\Statex\leavevmode\llap{\rule{\dimexpr\labelwidth+\labelsep}{\phaserulewidth}}\rule{\linewidth}{\phaserulewidth}}
\makeatother

\setphaserulewidth{.7pt}

\title{Automated Real-Time Inspection in Indoor and Outdoor 3D Environments with Cooperative Aerial Robots}
\author{Andreas Anastasiou,
        Angelos Zacharia,  
        Savvas Papaioannou,\\ 
        Panayiotis Kolios,
        Christos G. Panayiotou,
        and Marios M. Polycarpou
\thanks{The authors are with the KIOS Research and Innovation Center of Excellence (KIOS CoE) and the Department of Electrical and Computer Engineering, University of Cyprus, Nicosia, 1678, Cyprus.{\{\tt\small anastasiou.antreas, zacharia.angelos, papaioannou.savvas, pkolios, christosp, mpolycar\}@ucy.ac.cy}, (Andreas Anastasiou and Angelos Zacharia contributed equally to this work.)}}

\begin{document}

\maketitle
\thispagestyle{empty}
\pagestyle{empty}

\begin{abstract}
    This work introduces a cooperative inspection system designed to efficiently control and coordinate a team of distributed heterogeneous UAV agents for the inspection of 3D structures in cluttered, unknown spaces. Our proposed approach employs a two-stage innovative methodology. Initially, it leverages the complementary sensing capabilities of the robots to cooperatively map the unknown environment. It then generates optimized, collision-free inspection paths, thereby ensuring comprehensive coverage of the structure's surface area. The effectiveness of our system is demonstrated through qualitative and quantitative results from extensive Gazebo-based simulations that closely replicate real-world inspection scenarios, highlighting its ability to thoroughly inspect real-world-like 3D structures. 
\end{abstract}


\section{Introduction}
    Unmanned Aerial Vehicles (UAVs), have nowadays, revolutionized numerous fields by providing a versatile platform for a plethora of applications. Such applications include search and rescue \cite{terzi2019swifters,papaioannou2021towards,farooq2022uav,papaioannou2020coordinated}, search and track \cite{anastasiou2021hyperion,anastasiou2023sat,papaioannou2019decentralized}, security operations \cite{souli2020horizonblock,papaioannou2020cooperativeSecurity,papaioannou2021downing}, and infrastructure inspection \cite{savva2021icarus,zacharia2023distributed,papaioannou2023cooperative}. In this work we will focus on 3D infrastructure inspection which involves the examination of structures such as buildings, bridges, power lines, and wind turbines to detect any anomalies or damages. Traditional methods of inspection often involve manual labor, which can be time-consuming, costly, and potentially dangerous. Quadrotor UAVs, with their high maneuverability, can easily navigate and capture high-resolution images or videos, providing a safer and more efficient alternative. The use of UAVs for 3D infrastructure inspection involves equipping the drones with various sensors like cameras to capture detailed information about the structures. Advanced algorithms and techniques are then used to process and analyze this data, identifying potential issues that may require further investigation or immediate attention. 
    The recent advancements in aerial robotics, particularly UAVs, have significantly broadened the horizon of possibilities for using multiple drones in a coordinated manner for large-scale and complex 3D infrastructure inspections. This not only increases the efficiency of the inspection process but also reduces the time taken to successfully accomplish the task at hand. However, despite the promising benefits, there are still challenges to be addressed, such as navigating unknown indoor and outdoor environments with obstacles, inspecting complex structures, and the complexity of devising mechanisms for the real time coordination and efficient utilization of robots with heterogeneous capabilities.

    This work was motivated by the Cooperative Aerial Robots Inspection Challenge (CARIC) competition, held during the 2023 IEEE Conference on Decision and Control (CDC), whose purpose was to simulate and benchmark multi-UAV infrastructure inspection methodologies in real-world-like inspection scenarios. As such, in this paper, we propose a novel Cooperative Aerial Robots Inspection (CARI) methodology for inspecting complex 3D infrastructures in both indoor and outdoor environments. Our approach considers a system of multiple heterogeneous quadrotor UAVs each equipped with different sensors, such as gimballed cameras and LiDAR, each capable of inspecting the surface area of the infrastructure of interest using only limited knowledge about the environment, under communication and time constraints. To evaluate the quality of the inspection carried by the fleet of UAVs, both metrics for the level of bluriness and the level of resolution are employed assessing the observations of the camera sensors. The key contributions of the proposed scheme are as follows:
    \begin{itemize}
        \item We propose a cooperative inspection system designed for the efficient coordination and control of a distributed team of heterogeneous UAV agents assigned to inspect 3D structures in complex, cluttered indoor and outdoor environments. Our approach employs a two-stage strategy. Initially, it leverages the heterogeneous sensing capabilities of the UAVs to cooperatively map the unknown environment. Subsequently, it generates cooperative, collision-free inspection paths in real time, thereby ensuring full coverage of the structure's surface area.
        \item The effectiveness of our system is demonstrated through extensive qualitative and quantitative results from Gazebo-based simulations. These simulations closely replicate real-world inspection scenarios, underscoring the system's capability and practicality to thoroughly inspect 3D structures that closely resemble those found in the real-world settings. 
    \end{itemize}

    The rest of this paper is structured as follows. Section \ref{sec:relatedWork} delves into related work on infrastructure inspection. In Section \ref{sec:preliminaries}, we introduce the system model. Following this, the infrastructure inspection problem is introduced in Section \ref{sec:problemStatement}. Our proposed approach for online cooperative 3D inspection planning is presented in Section \ref{sec:proposedApproach}. Section \ref{sec:evaluation}  focuses on the evaluation of our approach, and finally Section \ref{sec:conclusions} concludes the paper.

\section{Related Work}
\label{sec:relatedWork}
    Infrastructure inspection is a critical part of ensuring the safety and functionality of essential facilities such as spanning bridges, pipelines, and other civil structures. Traditional inspection practices often pose challenges in terms of cost, safety, and efficiency. As technology advances, there is a growing exploration of robotic systems as a viable alternative for infrastructure inspection. These systems have the potential to mitigate risks associated with human inspectors, enhance data collection through advanced sensing technologies, and provide a cost-effective and efficient means of evaluating the structural integrity of vital infrastructure components. This shift towards robotic inspection solutions is underpinned by the need for innovations in mobility, autonomy, and sensing, as well as a growing emphasis on purpose-driven designs and multidisciplinary collaborations \cite{lattanzi2017review,lee2023survey}. 

    In the literature, numerous works explore the use of a single UAV for efficiently generating optimal paths to facilitate infrastructure inspection. More specifically, the study outlined in \cite{jing2016sampling} introduces a sampling-based approach for visual path planning. Designed for a camera-equipped UAV, this method visually covers a specific target of interest. Recasting the problem of coverage path planning (CPP) as a mixed-integer programming (MIP) problem has also been considered in the literature. The authors in \cite{papaioannou2023integrated} proposed a CPP algorithm for visually covering a 2D object of interest by formulating a MIP that utilizes a set of visibility constraints (i.e. camera-rays). Moreover, the authors in \cite{papaioannou20213d} address a 3D case of the CPP by recasting it as a MIP problem, enabling a single-UAV system to efficiently cover cuboid-like objects of interest. Our previous work, \cite{papaioannou2022uav}, introduces a UAV-based inspection planning approach utilizing a novel receding horizon MIP control strategy. The resulting paths achieve optimal inspection of feature points scattered on the surface of the target object. Furthermore, the authors in \cite{zhu2021online} propose a path planning methodology for information gathering utilizing a single-UAV system. The proposed approach involves conducting a discrete search by generating optimal paths, based on on a predetermined set of viewpoints, that maximize the information gathered based on recent observations. In \cite{jing2019coverage}, a framework has been introduced to address 3D coverage by directly sampling and planning paths within intricate 3D environments. The approach uses voxel dilation and subtraction based sampling to extract via-points and path primitives around the target. Subsequently, a primitive coverage graph is proposed that encodes the extracted information and finally graph search is used to calculate the inspection paths. In \cite{papachristos2016distributed} an algorithm for 3D inspection path planning for multiple structures using time-constrained aerial robots is presented. It optimizes paths by considering spatial properties and importance of infrastructure facilities, adhering to robot constraints, employing multi-step optimization. 
    
    Improving the path for a single UAV does not lead to a substantial decrease in inspection time when contrasted with employing multiple UAVs. The latter not only accelerates the inspection process but also strengthens the system resilience, by eliminating the single point of failure. To address the limitations of single-UAV systems, researchers have explored the use of multi-UAV approaches. This allows for the development of systems that can assign tasks to UAVs for cooperative infrastructure inspection, enhancing the reliability, accuracy, and speed of the inspection process. In \cite{jing2020multi}, a framework for multi-UAV Coverage Path Planning (CPP) is presented for inspecting large, complex 3D structures. This framework combines the Set Covering Problem and the Vehicle Routing Problem into a Set-Covering Vehicle Routing Problem, which is solved using a modified Biased Random Key Genetic Algorithm. \cite{mansouri2018cooperative} proposes a novel scheme to solve the CPP problem by slicing the infrastructure into horizontal planes and assigning areas to each UAV. \cite{hardouin2020next} introduces a path planning algorithm for multiple UAVs, featuring a Next-Best-View planner for surface reconstruction, using a greedy allocation of configurations and the Traveling Salesman Problem. It also employs a volumetric representation of surfaces, called Truncated Signed Distance Function, for cautious navigation and evaluation of 3D reconstruction. \cite{ivic2023multi} presents a unique path planning algorithm based on a potential field approach. This algorithm uses a refined version of the Heat Equation Driven Area Coverage technique to minimize the difference between desired and actual coverage. It also incorporates a camera direction control feature that ensures comprehensive coverage of the structure within the sensor’s field of view. Lastly, \cite{zacharia2023distributed} introduces an online trajectory planning algorithm for a multi-UAV system to thoroughly inspect large-scale complex structures in a 3D environment. This is achieved by utilizing a distributed control scheme for each UAV that operates based on local measurements.

\section{Preliminaries}
    \label{sec:preliminaries}
    In this work, we consider a fleet of heterogeneous quadrotor UAVs employed to inspect the surface of an infrastructure of interest, while avoiding potential static and dynamic obstacles in an unknown cluttered environment, under communication constraints. The fleet is composed of two types of UAVs: (a) explorer UAVs, equipped with rotating LiDAR and gimballed camera sensors, and (b) photographer UAVs, which are smaller and only carry a gimballed camera sensor. 

    \subsection{Operational Environment}
    \label{ssec:operational_environment}
    The operational environment, in which the UAVs are deployed, is an agnostic cluttered volume that needs to be mapped in real time to facilitate the identification of static obstacles and the infrastructure to be inspected. The infrastructure targeted for inspection can be represented by a triangular mesh. We assume the presence of a finite set $\mathcal{Z}\subset \mathbb{R}^3$ consisting of points of interest uniformly distributed across the surface of the infrastructure. These points must be discovered and observed, in real-time, by the UAV agents using their payload sensors. The quality of the visual inspection (i.e., how well a particular point $\zeta\in\mathcal{Z}$ is observed) is determined by a scoring scheme further described in \ref{ssec:sensors}.
    In this work, since we consider no prior map for the environment, the system operator restricts the operational volume by defining a set $\mathcal{B}$ of bounding boxes, thereby focusing the exploration to areas of high importance.

    \subsection{UAV Dynamical Model}
    \label{ssec:uav_model}
    We consider $N$ quadrotor UAVs, of which $N_e\in\{1,2\}$ are explorer UAVs and $N_p\in\{0,\dots,N-N_e\}$ are photographer UAVs, each free to move in the 3D Cartesian space. The inertial coordinate frame is defined in this space as $\mathbf{I} = \{x_{\mathbf{I}},y_{\mathbf{I}},z_{\mathbf{I}}\}$. The discretized dynamics of the quadrotor UAV agent can be expressed as:
    \begin{equation}
        \mathbf{x}_k = 
        \begin{bmatrix}
            \mathbb{I}_{4\times 4} & \delta t \cdot \mathbb{I}_{4\times 4} \\
            \mathbb{O}_{4\times 4} & \mathbb{I}_{4\times 4} 
        \end{bmatrix} \mathbf{x}_{k-1} +
        \begin{bmatrix}
            \mathbb{O}_{4\times 4} \\
             \delta t \cdot \mathbb{I}_{4\times 4}
        \end{bmatrix}
        \mathbf{u}_{k-1} 
        \label{eq:UAV_dynamics}
    \end{equation}
    where $\mathbf{x}_k = [\mathbf{p}_k, \psi_k, \mathbf{v}_k, \mathrm{v}_k^{\psi}]^T \in \mathbb{R}^8$ represents the state of the UAV agent at timestep $k$, comprising the position $\mathbf{p}_k \in \mathbb{R}^3$ and velocity $\mathbf{v}_k \in \mathbb{R}^3$, both in the inertial coordinate frame $\mathbf{I}$. The yaw angle and its angular velocity are denoted by $\psi_k$ and $\mathrm{v}_k^{\psi}$, respectively. Note that $\mathbb{I}_{4\times 4}$ and $\mathbb{O}_{4\times 4}$ represent the $4 \times 4$ identity and zero matrices, respectively. Additionally, $\delta t$ stands for the sampling interval. Moreover, $\mathbf{u}_k =[u_k^x,u_k^y,u_k^z,u_k^{\psi}]\in \mathbb{R}^4$ denotes the control input applied to the UAV, consisting of the linear and yaw acceleration in the inertial coordinate frame $\mathbf{I}$.

    \subsection{Ranging and Visual Sensor Models}
    \label{ssec:sensors}
    For efficiently reconstructing the unknown environment in real-time, the explorer UAVs are equipped with a LiDAR sensor, capable at providing dense measurements that represent the environment. More specifically, the measurement provided by the LiDAR sensor consists of a point-cloud drawn from the surrounding 3D space. Thus, the returned points lay on the surface of the nearest obstacles in the environment that are within the sensor's limited sensing range. This range is defined as a sphere-like shape around the sensor of radius $r_L$. The utilization of these measurements is crucial for obstacle avoidance and path planning purposes. 
    
    Furthermore, all UAV agents are equipped with gimballed camera sensors that possess rotation capabilities and are utilized for observing the interest points in $\mathcal{Z}$ with a certain quality. In this work, the camera view is represented as a rectangular pyramid, with its apex located at the camera's optical center. The pyramid is defined by the horizontal $\vartheta_h$ and vertical $\vartheta_v$  Field of View (FoV) angles, and the effective range $r_C$, with its volume is denoted as $\mathcal{C}$. On command, the FoV can be oriented in a specific direction by rotating the gimbal to the desired inclination angle $\theta \in [\theta_{\min}, \theta_{\max}]$ and azimuthal angle $\phi \in [\phi_{\min},\phi_{\max}]$, adhering to the gimbal's rotation constraints. The agent observes an interest point $\zeta$ within its camera FOV at a certain quality level (i.e., observation quality) based on the following observation metrics: (a) $q_{\zeta,\text{blur}} \in [0, 1]$, describing the level of blurriness with which the point was captured, and (b) $q_{\zeta,\text{res}} \in [0, 1]$, indicating the level of resolution. The level of bluriness is interpreted as the number of pixels that an interest point moves across during the exposure time $\tau$ and is derived from: 
    \begin{subequations}
        \begin{align}
            \label{eg:blur}
            &q_{\zeta,\text{blur}}=\min \left(\frac{c}{\max \left(\lVert u_1-u_0\rVert,\lVert v_1-v_0\rVert\right)}, 1.0\right) \\
            \label{eg:points}
            &u_0=f \cdot \frac{x_0}{z_0}, u_1=f \cdot \frac{x_1}{z_1}, v_0=f \cdot \frac{y_0}{z_0}, v_1=f \cdot \frac{y_1}{z_1} \\
            \label{eg:pointUpdate}
            &{\left[x_1, y_1, z_1\right]^{\top}=\left[x_0, y_0, z_0\right]^{\top}+\mathbf{v} \cdot \tau}
        \end{align}
    \end{subequations}
    where $c$ is the pixel width, $f$ is the focal length, $[x_0,y_0,z_0]^{\top}$ is the position of the captured point of interest, and $\mathbf{v} \in \mathbb{R}^2$ is the velocity of the interest point in the camera frame at the time of capture \cite{thien2023Determining}. Moreover, the resolution level is ensured to be satisfactory by assessing the estimated real-world size captured within a single image pixel (expressed in millimeters-per-pixel) and assigning a score based on:
    \begin{subequations}
        \begin{align}
            \label{eg:res}
            &q_{\zeta,\text{res}}=\min \left(\frac{r_{\text{des}}}{\max \left(r_{\text{horz}},r_{\text{vert}}\right)}, 1.0\right)\\
            \label{eg:horNvertResolutions}
            &r_{\text{horz}}=\frac{c}{\left\|u_4-u_3\right\|}\text{, } r_{\text{vert}}=\frac{c}{\left\|v_4-v_3\right\|}
        \end{align}
    \end{subequations}
    where $r_{\text{des}}$ is the desired resolution, and $\left(u_3, u_4\right)$ and $\left(v_3, v_4\right)$ represent the shifted positions of the interest point in $x$-axis and $y$-axis, respectively, in the camera frame.

    The point of interest $\zeta$ is observed at time-step $k$ when:
    \begin{equation}
        \zeta \in \bigcup_{i\in\{1,\ldots,N\}} \mathcal{C}_{i,k} \quad, \quad \quad q_{\zeta,k} > \bar{q}
    \end{equation}
    where, $q_{\zeta,k} = q_{\zeta,\text{blur},k} \cdot q_{\zeta,\text{res},k}$ is the observation quality of point $\zeta$, and $\mathcal{C}_{i,k}$ is the FoV of the $i_{\text{th}}$ UAV agent, both at time-step $k$. Parameter $\Bar{q}$ is a tuning parameter that describes the lower admissible value of observation quality for each point of interest.

    \subsection{Communication Model}
    \label{ssec:communication}
    The exchange of information between UAV agents occurs in a peer-to-peer fashion only when two agents are within Line of Sight (LoS) with each other. In other words, LoS means that there are no obstructions, such as buildings or other obstacles, blocking the direct view between two UAVs. If an obstruction is present, communication is not possible. The group of agents that agent $i$ can communicate with at a specific time-step $k$ is represented as $\mathcal{N}_{i,k}$, with $i \in \{1,\ldots,N\}$. The network topology can change over time due to agent movement, which may affect LoS. Therefore, in this work the agents opportunistically communicate whenever they are within LoS conditions and exchange information such as their current positions, and knowledge of the environment (i.e., their maps) to cooperatively map the environments, and coordinate the inspection task.

\section{Problem Statement}
    \label{sec:problemStatement}
    An infrastructure within an unknown cluttered environment has to be inspected with the maximum observation quality while adhering to a time limit constrain. A heterogeneous UAV fleet, governed by Eq. \eqref{eq:UAV_dynamics}, is employed to observe interest points across the infrastructure's surface using several payload sensors under communication limitations. The observation quality is assessed by the scoring metrics described in Sec. \ref{ssec:sensors}. The 3D infrastructure inspection problem tackled in this work can be stated as follows: \textit{Given a set of bounding boxes $\mathcal{B}$, a heterogeneous team of $N$ UAV agents ($i\in \{1,..,N\}$), and mission time $\mathcal{T}$, find a set of cooperative inspection paths (one for every agent) that maximizes the inspection score $\mathcal{Q}$ given by:}
    \begin{equation}
        \mathcal{Q} = \sum_{\zeta\in\mathcal{Z}} \max_{i \in \{1,\ldots,N\}} \left(\max_{k \in \{1,\ldots,\mathcal{T}\}} q_{\zeta,i,k} \right).
    \end{equation}
    This entails finding the joint kinematic and sensing control inputs of the agents that enable efficient and safe mapping of the environment while avoiding obstacles, discovering points of interest on the structure's surface, and observing them in a manner that maximizes observation quality and consequently the inspection score within the mission time.

\section{Proposed Approach}
    \label{sec:proposedApproach}
    \begin{figure}[t!]
        \centering
        \includegraphics[clip,trim=0cm 0cm 0cm 0cm,width=\columnwidth]{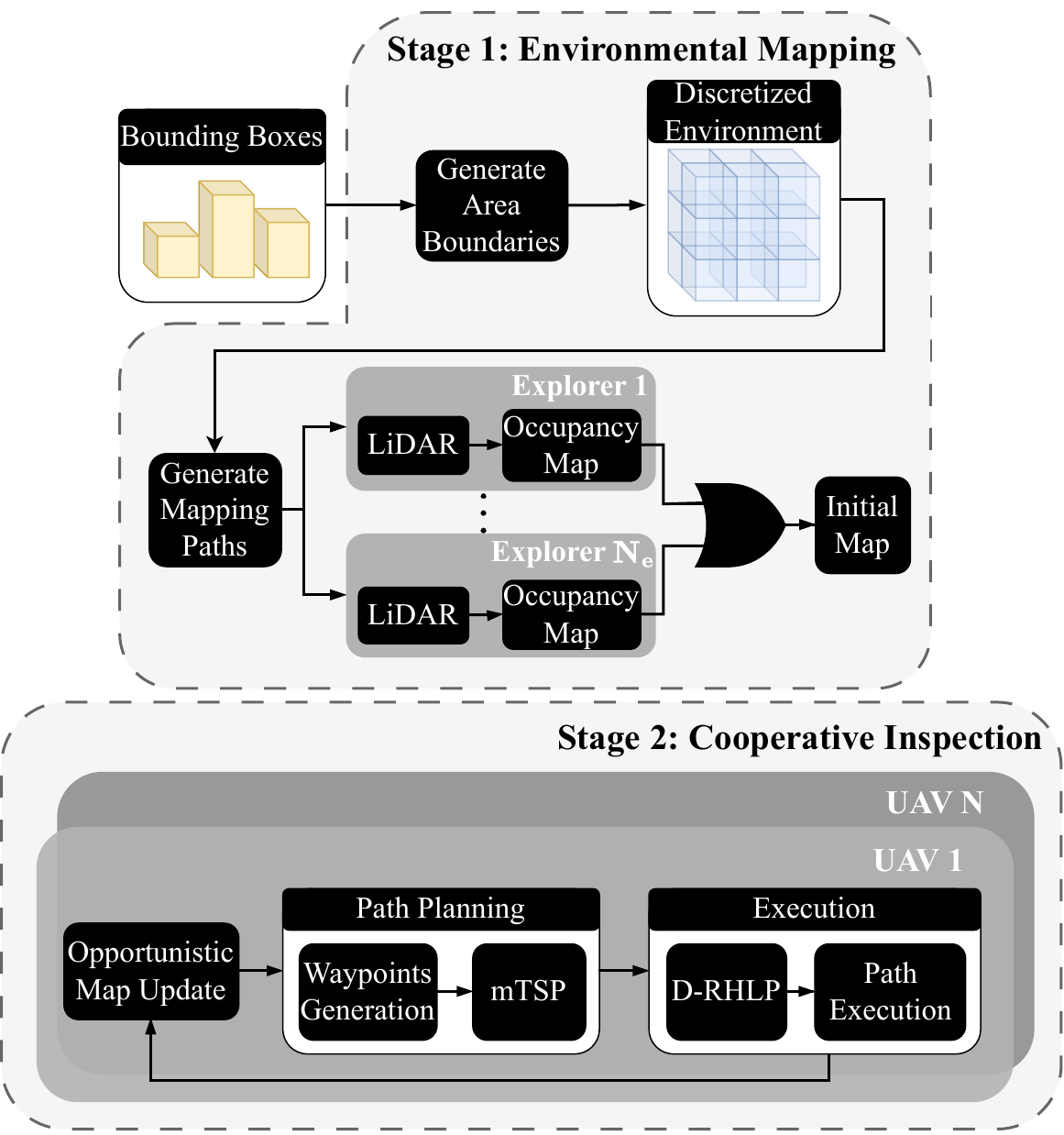}        
        \caption{Overview of the proposed approach for 3D infrastructure inspection using multi-UAV system.}
        \label{fig:overview}
    \end{figure}
    
    \begin{figure*}
        \centering
        \includegraphics[width=0.8\textwidth]{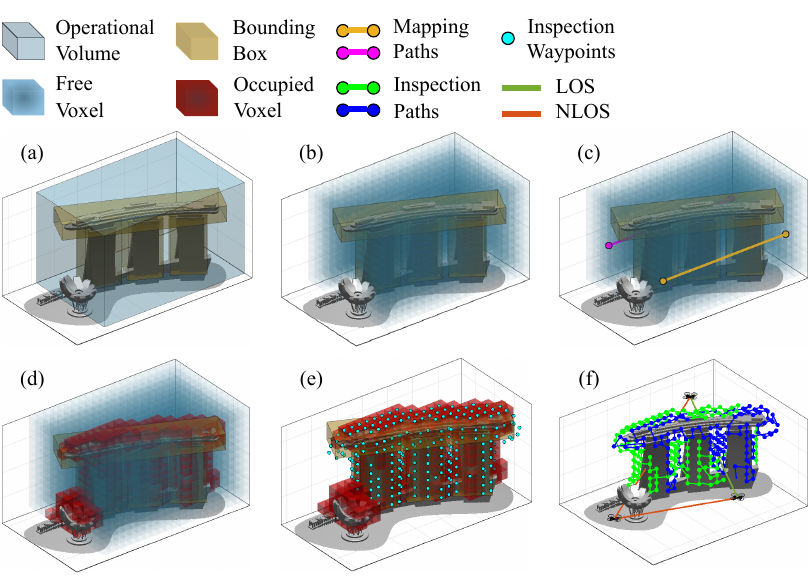} 
        \caption{(a) Derivation of the operational volume, (b) Discretization of the operational volume, (c) Mapping path generation and execution, (d) Initial occupancy map generation, (e) Inspection waypoint generation, (f) Inspection path generation.}
        \label{fig:proposedApproach}
    \end{figure*}

    \begin{algorithm}
        \caption{Proposed CARI Scheme}
        \label{alg:cari}
        \begin{algorithmic}[1]
            \Require A set of bounding boxes $\mathcal{B}$, enclosing the interest points in $\mathcal{Z}$, and the mission time $\mathcal{T}$.
            \Ensure Maximization of $\mathcal{Q}$.
                
            \phase{1}{Environmental Mapping}
                \State Retrieves $\mathbf{p}_{i,0} \in \mathcal{P}$, $i=\{1,\ldots,N\}$
                \State Calculates the boundaries of $\mathcal{O}$, as in $\eqref{eq:areaBoundaries}$
                \State Forms $\mathcal{G}$ by dividing $\mathcal{O}$ by the voxel size $\text{V}$.
                \State Generates $N_e$ mapping paths that traverse $\mathcal{O}$ 
                \For{\textbf{each} explorer $i\in\{1,\dots,N_e\}$}
                    \While{Executing mapping path}
                        \State Gathers LiDAR measurements 
                        \State Constructs and updates an occupancy map $\mathcal{M}_{i,k}$
                    \EndWhile
                    \State Shares $\mathcal{M}_{i,k}$ with $j$ UAV,  $j\in\mathcal{N}_{i,k}$
                    \State Receives $\mathcal{M}_{j,k}$, $\forall$ $j\in\mathcal{N}_{i,k}$
                    \State Creates an initial map $\mathcal{M}_{i,k}$ by merging the maps
                \EndFor
            \phase{2}{Cooperative Inspection}
                \For{\textbf{each} UAV $i\in\{1,\dots,N\}$}
                    \While{true}
                        \State Shares, receives and updates $\mathcal{M}_{i,k}$
                        \State Generates $w_{l,i,\kappa_i}$ and $\hat{n}_{l,i,\kappa_i}$, $l=\{1,\ldots,|\mathcal{W}_{i,\kappa_i}|\}$
                        \State Retrieves $\sigma_{i,\kappa_i}\subseteq\mathcal{W}_{i,\kappa_i}$ from Algorithm \ref{alg:tsp}
                        \For{\textbf{each} $w_{l,i,\kappa_i}\in\sigma_{i,\kappa_i}$}
                            \State Applies D-RHLP to reach $w_{l,i,\kappa_i}$
                            \State Rotates the camera based on $\hat{n}_{l,i,\kappa_i}$
                            \State Retrieves $q_{\zeta,i,k}$ for each $\zeta\in\mathcal{C}_{i,k}$
                        \EndFor
                        \State $\kappa_i \gets \kappa_i+1$
                    \EndWhile
                \EndFor
        \end{algorithmic}
    \end{algorithm}

    This section outlines our proposed methodology for inspecting 3D infrastructures using multiple UAVs. This work considers $N$ quadrotor UAVs, of which $N_e\in\{1,2\}$ are explorer UAVs and $N_p\in\{0,\dots,N-N_e\}$ are photographer UAVs, each one capable of inspecting the surface area $\mathcal{S}$ of the infrastructure of interest. The proposed CARI approach is divided into two key stages: (a) operational volume determination and environmental mapping, and (b) path generation for cooperative inspection. Figure \ref{fig:overview} depicts the overview of the two-stage approach. More specifically, at the initial stage an operational volume $\mathcal{O}$ is constructed with its boundaries calculated based on the set of bounding boxes $\mathcal{B}$ that enclose the set of interest points $\mathcal{Z}$ to be inspected. Stage one is also responsible for the environment mapping by the explorer UAVs generating an initial map of the environment. Finally, at stage two, the fleet of UAVs cooperatively generate and execute inspection paths in a manner that maximizes $\mathcal{Q}$. This process is performed in real-time in a closed-loop fashion until the end of the mission time $\mathcal{T}$. In the following sections, we will dive into the specifics of each component and discuss how these components interact and complement each other to create a comprehensive, robust, and scalable CARI system. This work operates on the assumption that each UAV possesses the capability to self-localize within the environment. The overall process is outlined in Algorithm \ref{alg:cari}.

    \subsection{Operational Volume Determination}
    The methodology proposed in this work starts with the determination of the area boundaries and the discretization resolution of the operational volume $\mathcal{O}$. As mentioned above, the only available information about the infrastructure to be inspected is a set of bounding boxes $\mathcal{B}$ that enclose the set of interest points $\mathcal{Z}$. The operational volume boundaries are thus calculated based on $\mathcal{B}$ and the initial positions of the UAV agents $\mathcal{P}=\{\mathbf{p}_{i,0}\}_{i=1}^N$, $\mathbf{p}_{i,0} \in \mathbb{R}^3$. Thus, Eq. \eqref{eq:areaBoundaries}, calculates the minimum and maximum values of each axis resulting in the creation of the smallest cuboid containing both the infrastructure to be inspected and the UAV fleet. A visual representation of this cuboid is shown in Fig. \ref{fig:proposedApproach}a.
    \begin{equation}
        \begin{aligned}
            & x\text{-axis: }\left\{
            \begin{array}{l}
                x_{\min } =\min \left(\min_x \mathcal{B}, \min_x \mathcal{P}\right) \\
                x_{\max }=\max \left(\max_x \mathcal{B}, \max_x \mathcal{P}\right)
            \end{array}\right. \\
            & y\text{-axis: }\left\{
            \begin{array}{l}
                y_{\min }=\min \left(\min_y \mathcal{B}, \min_y \mathcal{P}\right) \\
                y_{\max }=\max \left(\max_y \mathcal{B}, \max_y \mathcal{P}\right)
            \end{array}\right. \\
            & z\text{-axis: }\left\{
            \begin{array}{l}
                z_{\min }=\min \left(\min_z \mathcal{B}, \min_z \mathcal{P}\right) \\
                z_{\max }=\max \left(\max_z \mathcal{B}, \max_z \mathcal{P}\right)
            \end{array}\right.
        \end{aligned}
        \label{eq:areaBoundaries}
    \end{equation}
    As the area boundaries are defined, a 3D grid is formed by dividing $\mathcal{O}$ into voxels of size $\text{V} \times \text{V} \times \text{V}$ meters, as illustrated in Fig. \ref{fig:proposedApproach}b. The resulting grid is treated as a connected mesh graph $\mathcal{G} = (\mathcal{V}, \mathcal{E})$, where $\mathcal{E} = \{(u,v) \mid \Vert u-v \Vert < \text{V},$ $\forall$ $u,v \in \mathcal{V},$ $u\neq v\}$, with each voxel representing a vertex. Subsequently, each UAV agent in the fleet is aware of the adjacency matrix that describes $\mathcal{G}$.

    \subsection{Environmental Mapping}
    The second step in Stage 1 is crucial component in our proposed methodology. Its responsibility is to construct an initial map of $\mathcal{O}$. This stage is solely reliant on the explorer UAVs, which are equipped with LiDAR sensors. These sensors are capable of capturing detailed and accurate spatial data of the surrounding environment, providing a rich dataset for our mapping process. Upon the successful construction of $\mathcal{G}$ from the previous step, the mapping process is initiated with the generation of $N_e$ paths in a way that the explorer UAVs traverse $\mathcal{O}$ along its longest axis (see Fig. \ref{fig:proposedApproach}c). While executing the generated paths, each explorer utilizes measurements of the environment gathered by its LiDAR sensor. It independently constructs an occupancy map $\mathcal{M}_{i,k}$ by identifying the voxels in the 3D grid containing the 3D coordinates of the measurements (representing points of obstacles) and marking them as occupied. Once the paths are executed by the agents, they cooperatively integrate their individual occupancy maps by merging them, thereby creating an initial map of $\mathcal{O}$ shown in Fig. \ref{fig:proposedApproach}d. The initial map is then shared to every photographer UAV in the fleet opportunistically (i.e., when there is LoS with an explorer UAV). This initial map serves as the foundation for the subsequent stage of the infrastructure inspection process. It provides a preliminary layout of the operational area, allowing planning and executing the steps further needed for a successful and effective inspection. This is achieved by eliminating the inward edges of the corresponding occupied vertices in $\mathcal{G}$, thus preventing any path from traversing through an occupied voxel.

    \subsection{Cooperative Inspection}
    In the final stage of the inspection mission, each UAV takes advantage of any opportunity to engage in the exchange and update of its occupancy map $\mathcal{M}_{i,k}$ with every other UAV within its LoS. This process, is a dynamic one, allowing for real time updates and adjustments to the map based on the most recent data that the explorer UAVs continuously collect. Once the occupancy map is updated, each UAV $i$ then strategically generates a set of inspection waypoints $\mathcal{W}_{i,\kappa_i}$. These inspection waypoints are positioned around the occupied voxels that lie within the bounding boxes in $\mathcal{B}$, as depicted in Fig. \ref{fig:proposedApproach}e. Each waypoint $w_{l,i,\kappa_i}\in\mathcal{W}_{i,\kappa_i}$ is linked with a direction vector $\hat{n}_{l,i,\kappa_i}$, indicating the direction towards the occupied voxel that prompted the creation of the respective waypoint. This implies that the direction vector points towards the occupied voxel responsible for triggering the waypoint's generation. The generation of these inspection waypoints and direction vectors is a crucial step, setting the course for the UAV's path planning within the operational area and facilitating the calculation of the desired $\theta$ and $\phi$ gimbal angles to effectively inspect the infrastructure. 
    
    Following the generation of inspection waypoints, each UAV then addresses the multi-Traveling Salesman Problem (mTSP). This problem is solved cooperatively with other UAVs within LoS. The mTSP is a complex optimization problem that aims to determine the most efficient route for each UAV, given multiple destinations. The heuristic algorithm used to solve the mTSP in this work, is inspired by the proposed algorithm in \cite{anastasiou2020swarm}. It's crucial to understand that due to the communication constraints the mTSP is solved in a distributed manner. To elaborate, each UAV $i$ generates inspection paths $\sigma_{\kappa_i}$ considering the positions $\mathbf{p}_{j,{\kappa_i}}$, $j \in \mathcal{N}_{i,{\kappa_i}}$ and the generated inspection waypoints $\mathcal{W}_{i,{\kappa_i}}$. Consequently, UAV $i$ selects the inspection path $\sigma_{i,\kappa_i}$ specifically tailored to it, ensuring efficient and coordinated movement of the UAVs. A breakdown of the mTSP algorithm is given in Algorithm \ref{alg:tsp} and a visualization of its output is shown in Fig. \ref{fig:proposedApproach}f. It is important to note that, the inspection waypoints $\mathcal{W}_{i,\kappa_i}$, direction vectors $\hat{n}_{l,i,\kappa_i}$ and inspection paths $\sigma_{\kappa_i}$ of each UAV agent $i$, are regenerated every time the UAV agent successfully visits every $w_{l,i,\kappa_i} \in \sigma_{i,\kappa_i}$ (i.e., $\kappa_i = \kappa_i + 1$).
    
    \begin{algorithm}[t]
        \caption{Multiple Traveling Salesman Problem}
        \begin{algorithmic}[1]
            \Require A set $\mathcal{W}_{i,\kappa_i}$ containing the inspection waypoints and positions $\mathbf{p}_{j,k}$, $j\in \{i\cup\mathcal{N}_{i,k}\}$ of the UAVs that are within LoS with UAV $i$, at step $\kappa_i$.
            \Ensure $\sigma_{\kappa_i}$, s.t.: $\cup_{j=0}^{\{i \cup \mathcal{N}_{i,k}\}} \sigma_{j,\kappa_i} = \mathcal{W}_{i,\kappa_i}$.
            \State $\mathrm{v}_l \gets 0$ $\forall$ $l \in \{1,\dots\, \rvert\mathcal{W}_{i,\kappa_i}\lvert\}$
            \State $\sigma_{j,\kappa_i} \gets \{\mathbf{p}_{j,k}\}$ $\forall$ $j\in \{i\cup\mathcal{N}_{i,k}\}$
            \While{$\sum_{l=1}^{\rvert\mathcal{W}_{i,\kappa_i}\lvert} \mathrm{v}_l < \rvert\mathcal{W}_{i,\kappa_i}\lvert$}
                \For {$j\in \{i\cup\mathcal{N}_{i,k}\}$}
                    \State $s \gets$ last waypoint in path $\sigma_{j,\kappa_i}$
                    \State $t \gets$ closest $w_{l,i,\kappa_i} \in \mathcal{W}_{i,\kappa_i}$ from $s$ s.t.: $\mathrm{v}_l=0$
                    \State $\sigma_{j,\kappa_i} \gets  \sigma_{j,\kappa_i} \cup t$
                    \State $\mathrm{v}_l \gets 1$
                \EndFor
            \EndWhile
        \end{algorithmic}
        \label{alg:tsp}
    \end{algorithm}

    To execute the inspection path $\sigma_{i,\kappa_i}$ generated by the cooperative mTSP solution, each UAV $i$ employs a strategy we reffered to as Dijkstra-Receding Horizon Local Planning (D-RHLP). D-RHLP generates at each time-step $k$ an optimal local collision-free path, based on the current state of $\mathcal{M}_{i,k}$, starting from $\mathbf{p}_{i,k}$ to the receding $w_{l,i,\kappa_i}\in\sigma_{i,\kappa_i}$ by exploiting Dijkstra's algorithm \cite{dijkstra1959note}. It executes a segment of the generated local path and then goes through a cycle of replanning and execution until the UAV reaches the desired waypoint. More specifically, the receding horizon technique is utilized due to the dynamic nature of the environment. Addressing inter-agent collision avoidance, all the voxels within which the UAVs are currently located are marked as occupied, ensuring that no two UAVs coexist in the same voxel. The adjacency matrix, given as input to Dijkstra's algorithm, represents the graph $\mathcal{G}$. In this graph, each vertex communicates exclusively with its immediate voxel-sized hop neighbor vertices. Note that each voxel previously marked as occupied does not communicate with any other voxel. While UAV $i$ executes its local plan, provided by D-RHLP, it rotates its camera's FoV to point towards the direction described by $\hat{n}_{l,i,\kappa_i}$, which is associated with the last waypoint in the local plan, i.e. the receding waypoint $w_{l,i,\kappa_i}$.
    
    This comprehensive process is iteratively repeated until the conclusion of the mission time. The iterative nature of the procedure is further emphasized by the dynamic exchange of occupancy maps whenever two UAVs come within LoS. This continuous exchange of information allows for real time updates and refinements to the mission strategy. Based on the updated occupancy map resulting from this exchange, each UAV regenerates and refines its set of inspection waypoints $\mathcal{W}_{i,\kappa_i}$. It then cooperatively solves the mTSP problem with neighboring UAVs once again. The refined plan, which represents the most efficient path given the current information, is then executed. This cyclic process of exchange, update, regeneration, and execution persists until the mission’s conclusion. It ensures that the UAVs are always operating based on the most recent and accurate information, thereby increasing the efficiency and success rate of the mission.

\section{Evaluation}
    \label{sec:evaluation}
    \begin{figure*}[t]
        \centering
        \includegraphics[width=\textwidth]{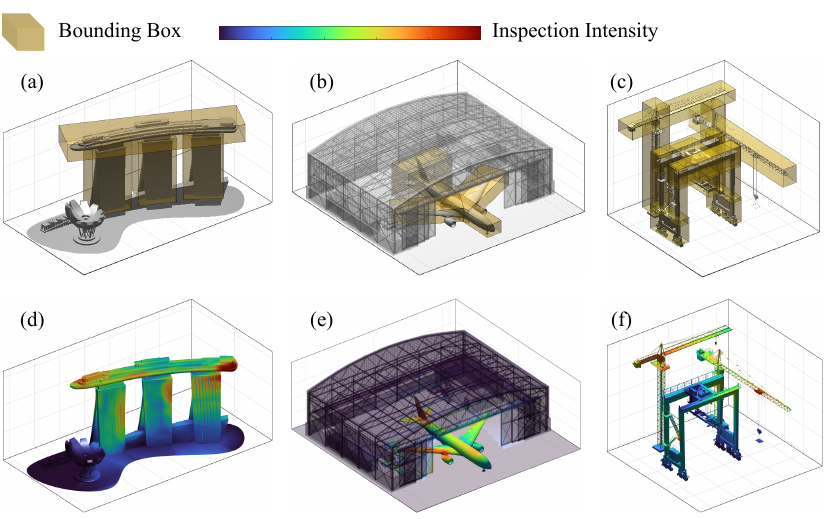} 
        \caption{(a)-(c) Illustrative examples of the 3 inspection scenarios conducted for the evaluation of the proposed approach. The bounded boxes indicate the selected area for inspection. (d)-(f) Visualization of inspection intensity achieved by the UAV fleet at each scenario, respectively, by employing the proposed CARI scheme.}
        \label{fig:results}
    \end{figure*}

    \begin{figure}
        \centering
        \includegraphics[clip, trim= 3.4cm 8cm 4.3cm 9cm, width=\columnwidth]{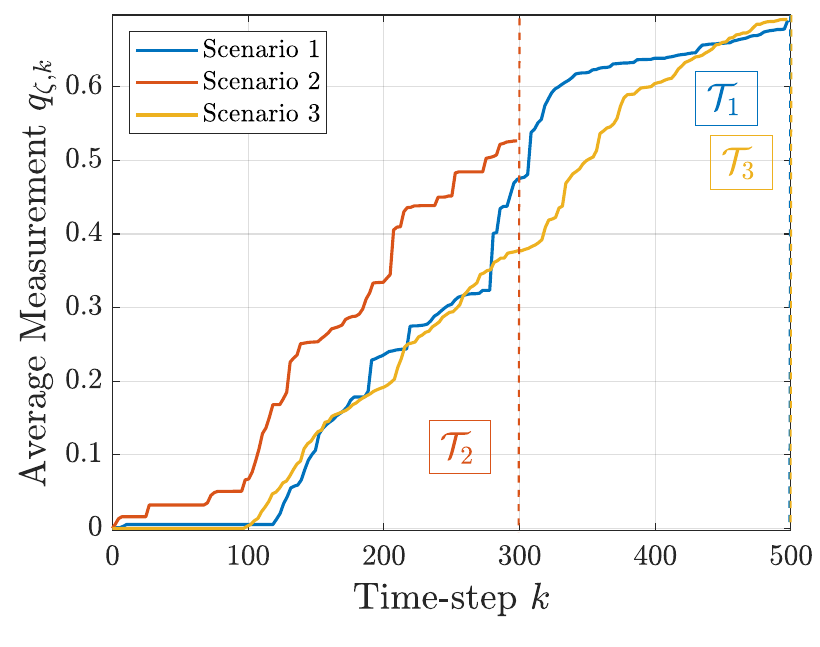} 
        \caption{Average observation quality score $q_{\zeta,k}$ achieved up to time-step $k$ for each scenario for inspection mission durations of $\mathcal{T}_{1} = 300$, $\mathcal{T}_{2} = 500$, and $\mathcal{T}_{3} = 500$, respectively.}
        \label{fig:score}
    \end{figure}

    Extensive simulations have been conducted to evaluate the effectiveness of the proposed approach in several real-word-like infrastructure inspection scenarios, in a Gazebo-based environment.

    \subsection{Simulation Setup}
    
    \begin{table}
        \centering
        \caption{Scenario Setups}
        \setlength{\tabcolsep}{6pt} 
        \renewcommand{\arraystretch}{1.2} 
        \resizebox{\columnwidth}{!}{%
        \begin{tabular}{|c|c|c|c|c|c|} \hline
             \textbf{Scenario} &
             \textbf{\begin{tabular}[c]{@{}c@{}} Infrastructure\\ Size $[m]$\end{tabular}} &
             \textbf{\begin{tabular}[c]{@{}c@{}}Interest\\ Points $|\mathcal{Z}|$\end{tabular}} & 
             \textbf{\begin{tabular}[c]{@{}c@{}}Mission\\ Time $\mathcal{T}$ $[s]$\end{tabular}} & 
             \textbf{\begin{tabular}[c]{@{}c@{}}Explorer\\ UAVs $N_e$\end{tabular}} & 
             \textbf{\begin{tabular}[c]{@{}c@{}}Photographer\\ UAVs $N_p$\end{tabular}}\\ \hline
            1& 140 $\times$ 60 $\times$ 60 & 3973 & 500 & 2 & 3 \\ \hline
            2& 120 $\times$ 80 $\times$ 40 & 1615 & 300 & 1 & 2\\ \hline
            3& 100 $\times$ 80 $\times$ 80 & 4935 & 500 & 2 & 3\\ \hline
        \end{tabular}
        }
        \label{tab:scenarios}
    \end{table}

    The evaluation setup is an open-source scheme based on the Gazebo Simulator, RotorS, and the Robot Operating System (ROS). The Gazebo simulator is responsible for rendering the three given real-world-like infrastructure scenarios including, a large building structure (i.e., a reconstruction of the Marina Bay Sands hotel in Singapore), a 3D reconstructed airplane residing inside its hangar, and a crane inspection scenario, as shown in Fig. \ref{fig:results}a,b and c, respectively. Each 3D object in the three infrastructure scenarios is represented by a triangular mesh surface $\mathcal{S}$. In each scenario, the set of interest points $\mathcal{Z}$ to be inspected were scattered along the surface of the infrastructure. More details about each scenario including size and number of interest points are shown in Table \ref{tab:scenarios}. As mentioned before, the environment is represented as an occupancy map which is generated by utilizing the OctoMap, a 3D mapping framework \cite{hornung2013octomap}. The voxel size of the occupancy map is selected as $\text{V} = 6$ $[m]$ The models of the two types of quadrotor UAVs are provided by the RotorS simulator which is responsible for simulating the flight dynamics and inertial sensors, ensuring accurate localization. Both types are equipped with a gimballed camera sensor capable of capturing information within its limited FoV, described by $\vartheta_v = 60^{\circ}$ and $\vartheta_h = 80^{\circ}$. Moreover, the gimbal to which the camera sensor is attached, is constraint by  $\theta\in[-90^{\circ}, 80^{\circ}]$ and $\phi\in[-90^{\circ}, 90^{\circ}]$. Despite their similarity in regards to the camera sensor, the explorer UAVs are double both in size and weight. Taking advantage of their greater payload capacity, explorer UAVs also carry a Velodyne VLP-16 LiDAR sensor capable of gathering environmental information up to $50$ meters away with zero noise. In addition, the LiDAR is attached on a rotating servo mechanism that shifts the LiDAR's beams around the body-frame $x$-axis in the interval of $[-90^{\circ}, 90^{\circ}]$ completing a full cycle every $8$ seconds. Note that the number of deployed explorer and photographer UAVs, inspecting the infrastructure of interest within the mission time, differs in each scenario, as shown in Table \ref{tab:scenarios}. The exchange of information between the various components is achieved by the utilization of the ROS framework, acting as a middleware. The simulation setup as described above was provided as part of the CARIC competition by the organizing committee. 

    To ensure that our approach is practical and can be readily used in real-world applications, we have implemented it as a package for ROS. The implementation of our methodology was done using the Python programming language. Our implementation was developed and tested on the Nvidia Jetson AGX Orin Developer Kit. The Orin System-on-Chip integrates 12 ARM Cortex-A78AE CPU cores running at 2.2 GHz, and an Ampere GPU with 2048 CUDA cores and 64 Tensor cores running at 1.3 GHz as well as 64GB of LPDDR5 RAM. This makes it an ideal platform for compute-intensive applications like ours.

    \subsection{Simulation Results}
    To evaluate the effectiveness and efficiency of the proposed CARI methodology, we employ two Key Performance Indicators (KPIs). The primary KPI in this study is the total inspection intensity of the infrastructure achieved by the UAV fleet. This KPI offers a more comprehensive method for assessing the distribution of inspection over the surface of the triangular mesh $\mathcal{S}$. The secondary KPI is the average score of observation $q_{\zeta}$. This latter KPI offers insights into the quality of observations of each interest point. 

    For the inspection task of all three scenarios, we employ the proposed CARI scheme given in Algorithm \ref{alg:cari}. The given sets of bounding boxes $\mathcal{B}_m$, as shown in Fig. \ref{fig:results}a-c, serve as input for each scenario $m=\{1,2,3\}$. This results in a set of cooperative, collision-free inspection paths for all UAV agents, as well as direction vectors for the gimballed cameras carried by the UAV agents. Based on the algorithm's output, a heat-map is generated that describes the inspection coverage intensity. In this heat-map, the most inspected area is represented with a red color, whereas the least inspected area is represented by blue. Thus, the inspection intensity achieved by employing the proposed CARI scheme for each scenario is illustrated in Fig. \ref{fig:results}d-f. It can be observed that the surfaces encapsulated within the given bounding boxes have been successfully inspected by the fleet, without wasting resources inspecting parts of the infrastructure that lie outside of the bounding boxes. This results into a relatively uniform inspection intensity which is desired since the interest points lay uniformly across the surface of the infrastructure. However, some parts exhibit higher intensity. This is mainly observed at transitional areas where the UAVs reduced speed, either for drastically changing course or for collision avoidance purposes.

    Beyond the inspection intensity, the quality of the observation is another crucial factor. For instance, covering the surface of an infrastructure at a higher speed, or lower resolution, might result in complete and uniform coverage, but lead to non-informative inspection. Therefore, the average observation quality for each scenario at each time-step was also recorded and is depicted in Fig. \ref{fig:score}. It can be observed that all three scenarios show an increasing trend in the average observation over time. This suggests that the average observation $q_{\zeta}$ improves as the time-step $k$ progresses, demonstrating that the proposed scheme achieves high levels of inspection quality regardless of the operating environment.

    The proposed CARI system, has participated in the Cooperative Aerial Robots Inspection Challenge of the 62nd IEEE Conference on Decision and Control (CDC), held in 2023 in Singapore. The effectiveness and practicality of the proposed approach was proven by taking the First place (i.e., ranking 1st), among 19 international teams. The total score for all three scenarios of the top five out of ten trials from the CARIC evaluation are provided in Table \ref{tab:caric}. A video demonstration of our proposed approach can be accessed at \url{https://youtu.be/pJtwYsDdd2U}.


    \begin{table}[]
        \centering
        \caption{CARIC Competition Results}
        \label{tab:caric}
        \setlength{\tabcolsep}{6pt} 
        \renewcommand{\arraystretch}{1.2} 
        \resizebox{\columnwidth}{!}{%
        \begin{tabular}{|c|c|c|c|c|c|}
        \hline
        \textbf{Trial} & 1       & 2       & 3       & 4       & 5 \\ \hline
        \textbf{Score} & 4245.35 & \textbf{4559.80} & 4241.84 & 3933.50 & 3724.61 \\ \hline
        \end{tabular}%
        }
    \end{table}

\section{Conclusions}
\label{sec:conclusions}
    In conclusion, this work has presented a novel CARI scheme that leverages the advanced mobility of aerial robots for efficient infrastructure inspection. The proposed scheme addressed the challenge of 3D path planning in diverse environments by employing multiple heterogeneous autonomous UAV agents for cooperative mapping and inspection under limited sensing capabilities and communication constraints. The effectiveness of our two-stage approach has been substantiated through extensive real-world-like simulations, showcasing its potential for practical 3D infrastructure inspection.
    




\section*{Acknowledgment}
This work was undertaken as part of the GLIMPSE project EXCELLENCE/0421/0586 which is co-financed by the European Regional Development Fund and the Republic of Cyprus through the Research and Innovation Foundation’s RESTART 2016-2020 Programme for Research, Technological Development and Innovation and supported by the European Union’s Horizon 2020 research and innovation programme under grant agreement No 739551 (KIOS CoE), and from the Government of the Republic of Cyprus through the Cyprus Deputy Ministry of Research, Innovation and Digital Policy.

\bibliographystyle{IEEEtran}
\balance
\bibliography{references}

\begin{thebibliography}{10}
\providecommand{\url}[1]{#1}
\csname url@samestyle\endcsname
\providecommand{\newblock}{\relax}
\providecommand{\bibinfo}[2]{#2}
\providecommand{\BIBentrySTDinterwordspacing}{\spaceskip=0pt\relax}
\providecommand{\BIBentryALTinterwordstretchfactor}{4}
\providecommand{\BIBentryALTinterwordspacing}{\spaceskip=\fontdimen2\font plus
\BIBentryALTinterwordstretchfactor\fontdimen3\font minus
  \fontdimen4\font\relax}
\providecommand{\BIBforeignlanguage}[2]{{%
\expandafter\ifx\csname l@#1\endcsname\relax
\typeout{** WARNING: IEEEtran.bst: No hyphenation pattern has been}%
\typeout{** loaded for the language `#1'. Using the pattern for}%
\typeout{** the default language instead.}%
\else
\language=\csname l@#1\endcsname
\fi
#2}}
\providecommand{\BIBdecl}{\relax}
\BIBdecl

\bibitem{terzi2019swifters}
M.~Terzi, A.~Anastasiou, P.~Kolios, C.~Panayiotou, and T.~Theocharides,
  ``{SWIFTERS}: A multi-{UAV} platform for disaster management,'' in \emph{2019
  International Conference on Information and Communication Technologies for
  Disaster Management (ICT-DM)}, 2019, pp. 1--7.

\bibitem{papaioannou2021towards}
S.~Papaioannou, P.~Kolios, T.~Theocharides, C.~G. Panayiotou, and M.~M.
  Polycarpou, ``Towards automated {3D} search planning for emergency response
  missions,'' \emph{Journal of Intelligent \& Robotic Systems}, vol. 103,
  no.~1, p.~2, 2021.

\bibitem{farooq2022uav}
A.~Farooq, A.~Anastasiou, N.~Souli, C.~Laoudias, P.~S. Kolios, and
  T.~Theocharides, ``Uav autonomous indoor exploration and mapping for {SAR}
  missions: Reflections from the {ICUAS} 2022 competition,'' in \emph{2022 19th
  International Conference on Ubiquitous Robots (UR)}, 2022, pp. 621--626.

\bibitem{papaioannou2020coordinated}
S.~Papaioannou, S.~Kim, C.~Laoudias, P.~Kolios, S.~Kim, T.~Theocharides,
  C.~Panayiotou, and M.~Polycarpou, ``Coordinated crlb-based control for
  tracking multiple first responders in 3d environments,'' in \emph{2020
  International Conference on Unmanned Aircraft Systems (ICUAS)}, 2020, pp.
  1475--1484.

\bibitem{anastasiou2021hyperion}
A.~Anastasiou, R.~Makrigiorgis, P.~Kolios, and C.~Panayiotou, ``Hyperion: A
  robust drone-based target tracking system,'' in \emph{2021 International
  Conference on Unmanned Aircraft Systems (ICUAS)}, 2021, pp. 927--933.

\bibitem{anastasiou2023sat}
A.~Anastasiou, S.~Papaioannou, P.~Kolios, and C.~G. Panayiotou, ``Model
  predictive control for multiple castaway tracking with an autonomous aerial
  agent,'' in \emph{2023 European Control Conference (ECC)}, 2023, pp. 1--8.

\bibitem{papaioannou2019decentralized}
S.~Papaioannou, P.~Kolios, T.~Theocharides, C.~G. Panayiotou, and M.~M.
  Polycarpou, ``Decentralized search and track with multiple autonomous
  agents,'' in \emph{2019 IEEE 58th Conference on Decision and Control (CDC)},
  2019, pp. 909--915.

\bibitem{souli2020horizonblock}
N.~Souli, R.~Makrigiorgis, A.~Anastasiou, A.~Zacharia, P.~Petrides, A.~Lazanas,
  P.~Valianti, P.~Kolios, and G.~Ellinas, ``{HorizonBlock}: Implementation of
  an autonomous counter-drone system,'' in \emph{2020 International Conference
  on Unmanned Aircraft Systems (ICUAS)}.\hskip 1em plus 0.5em minus 0.4em\relax
  IEEE, 2020, pp. 398--404.

\bibitem{papaioannou2020cooperativeSecurity}
S.~Papaioannou, P.~Kolios, C.~G. Panayiotou, and M.~M. Polycarpou,
  ``Cooperative simultaneous tracking and jamming for disabling a rogue
  drone,'' in \emph{2020 IEEE/RSJ International Conference on Intelligent
  Robots and Systems (IROS)}.\hskip 1em plus 0.5em minus 0.4em\relax IEEE,
  2020, pp. 7919--7926.

\bibitem{papaioannou2021downing}
S.~Papaioannou, P.~Kolios, and G.~Ellinas, ``Downing a rogue drone with a team
  of aerial radio signal jammers,'' in \emph{2021 IEEE/RSJ International
  Conference on Intelligent Robots and Systems (IROS)}, 2021, pp. 2555--2562.

\bibitem{savva2021icarus}
A.~Savva, A.~Zacharia, R.~Makrigiorgis, A.~Anastasiou, C.~Kyrkou, P.~Kolios,
  C.~Panayiotou, and T.~Theocharides, ``{ICARUS}: automatic autonomous power
  infrastructure inspection with {UAV}s,'' in \emph{2021 International
  Conference on Unmanned Aircraft Systems (ICUAS)}.\hskip 1em plus 0.5em minus
  0.4em\relax IEEE, 2021, pp. 918--926.

\bibitem{zacharia2023distributed}
A.~Zacharia, S.~Papaioannou, P.~Kolios, and C.~Panayiotou, ``Distributed
  control for {3D} inspection using multi-{UAV} systems,'' in \emph{2023 31st
  Mediterranean Conference on Control and Automation (MED)}.\hskip 1em plus
  0.5em minus 0.4em\relax IEEE, 2023, pp. 164--169.

\bibitem{papaioannou2023cooperative}
S.~Papaioannou, P.~Kolios, T.~Theocharides, C.~G. Panayiotou, and M.~M.
  Polycarpou, ``Cooperative receding horizon 3d coverage control with a team of
  networked aerial agents,'' in \emph{2023 62nd IEEE Conference on Decision and
  Control (CDC)}, 2023, pp. 4399--4404.

\bibitem{lattanzi2017review}
D.~Lattanzi and G.~Miller, ``Review of robotic infrastructure inspection
  systems,'' \emph{Journal of Infrastructure Systems}, vol.~23, no.~3, p.
  04017004, 2017.

\bibitem{lee2023survey}
A.~J. Lee, W.~Song, B.~Yu, D.~Choi, C.~Tirtawardhana, and H.~Myung, ``Survey of
  robotics technologies for civil infrastructure inspection,'' \emph{Journal of
  Infrastructure Intelligence and Resilience}, vol.~2, no.~1, p. 100018, 2023.

\bibitem{jing2016sampling}
W.~Jing, J.~Polden, W.~Lin, and K.~Shimada, ``Sampling-based view planning for
  {3D} visual coverage task with unmanned aerial vehicle,'' in \emph{2016
  IEEE/RSJ International Conference on Intelligent Robots and Systems
  (IROS)}.\hskip 1em plus 0.5em minus 0.4em\relax IEEE, 2016, pp. 1808--1815.

\bibitem{papaioannou2023integrated}
S.~Papaioannou, P.~Kolios, T.~Theocharides, C.~G. Panayiotou, and M.~M.
  Polycarpou, ``Integrated guidance and gimbal control for coverage planning
  with visibility constraints,'' \emph{IEEE Transactions on Aerospace and
  Electronic Systems}, vol.~59, no.~2, pp. 1276--1291, 2023.

\bibitem{papaioannou20213d}
------, ``{3D} trajectory planning for {UAV}-based search missions: An
  integrated assessment and search planning approach,'' in \emph{2021
  International Conference on Unmanned Aircraft Systems (ICUAS)}.\hskip 1em
  plus 0.5em minus 0.4em\relax IEEE, 2021, pp. 517--526.

\bibitem{papaioannou2022uav}
------, ``{UAV}-based receding horizon control for {3D} inspection planning,''
  in \emph{2022 International Conference on Unmanned Aircraft Systems (ICUAS)},
  2022, pp. 1121--1130.

\bibitem{zhu2021online}
H.~Zhu, J.~J. Chung, N.~R. Lawrance, R.~Siegwart, and J.~Alonso-Mora, ``Online
  informative path planning for active information gathering of a {3D}
  surface,'' in \emph{2021 IEEE International Conference on Robotics and
  Automation (ICRA)}.\hskip 1em plus 0.5em minus 0.4em\relax IEEE, 2021, pp.
  1488--1494.

\bibitem{jing2019coverage}
W.~Jing, D.~Deng, Z.~Xiao, Y.~Liu, and K.~Shimada, ``Coverage path planning
  using path primitive sampling and primitive coverage graph for visual
  inspection,'' in \emph{2019 IEEE/RSJ International Conference on Intelligent
  Robots and Systems (IROS)}.\hskip 1em plus 0.5em minus 0.4em\relax IEEE,
  2019, pp. 1472--1479.

\bibitem{papachristos2016distributed}
C.~Papachristos, K.~Alexis, L.~R.~G. Carrillo, and A.~Tzes, ``Distributed
  infrastructure inspection path planning for aerial robotics subject to time
  constraints,'' in \emph{2016 International Conference on Unmanned Aircraft
  Systems (ICUAS)}, 2016, pp. 406--412.

\bibitem{jing2020multi}
W.~Jing, D.~Deng, Y.~Wu, and K.~Shimada, ``Multi-{UAV} coverage path planning
  for the inspection of large and complex structures,'' in \emph{2020 IEEE/RSJ
  International Conference on Intelligent Robots and Systems (IROS)}.\hskip 1em
  plus 0.5em minus 0.4em\relax IEEE, 2020, pp. 1480--1486.

\bibitem{mansouri2018cooperative}
S.~S. Mansouri, C.~Kanellakis, E.~Fresk, D.~Kominiak, and G.~Nikolakopoulos,
  ``Cooperative coverage path planning for visual inspection,'' \emph{Control
  Engineering Practice}, vol.~74, pp. 118--131, 2018.

\bibitem{hardouin2020next}
G.~Hardouin, J.~Moras, F.~Morbidi, J.~Marzat, and E.~M. Mouaddib,
  ``{Next-Best-View planning for surface reconstruction of large-scale {3D}
  environments with multiple {UAV}s},'' in \emph{2020 IEEE/RSJ International
  Conference on Intelligent Robots and Systems (IROS)}.\hskip 1em plus 0.5em
  minus 0.4em\relax IEEE, 2020, pp. 1567--1574.

\bibitem{ivic2023multi}
S.~Ivi{\'c}, B.~Crnkovi{\'c}, L.~Grb{\v{c}}i{\'c}, and L.~Matlekovi{\'c},
  ``Multi-{UAV} trajectory planning for {3D} visual inspection of complex
  structures,'' \emph{Automation in Construction}, vol. 147, p. 104709, 2023.

\bibitem{thien2023Determining}
T.-M. Nguyen, ``{D}etermining the {V}elocity of a {W}orld {P}oint in a {M}oving
  {C}amera {F}rame,''
  \url{https://ntu-aris.github.io/caric/docs/CARIC_motion_blur.pdf}, 2023,
  [Accessed 07-02-2024].

\bibitem{anastasiou2020swarm}
A.~Anastasiou, P.~Kolios, C.~Panayiotou, and K.~Papadaki, ``Swarm path planning
  for the deployment of drones in emergency response missions,'' in \emph{2020
  International Conference on Unmanned Aircraft Systems (ICUAS)}.\hskip 1em
  plus 0.5em minus 0.4em\relax IEEE, 2020, pp. 456--465.

\bibitem{dijkstra1959note}
E.~W. Dijkstra, ``A note on two problems in connexion with graphs,''
  \emph{Numerische mathematik}, vol.~1, no.~1, pp. 269--271, 1959.

\bibitem{hornung2013octomap}
A.~Hornung, K.~M. Wurm, M.~Bennewitz, C.~Stachniss, and W.~Burgard, ``Octomap:
  An efficient probabilistic 3d mapping framework based on octrees,''
  \emph{Autonomous robots}, vol.~34, pp. 189--206, 2013.

\end{thebibliography}

\end{document}